\documentclass[pdflatex,sn-mathphys-num]{sn-jnl}

\usepackage{graphicx}%
\usepackage{multirow}%
\usepackage{amsmath,amssymb,amsfonts}%
\usepackage{amsthm}%
\usepackage{mathrsfs}%
\usepackage[title]{appendix}%
\usepackage{xcolor}%
\usepackage{textcomp}%
\usepackage{manyfoot}%
\usepackage{booktabs}%
\usepackage{algorithm}%
\usepackage{algorithmicx}%
\usepackage{algpseudocode}%
\usepackage{listings}%

\usepackage{float}
\usepackage{algpseudocode}
\usepackage{subcaption}

\theoremstyle{thmstyleone}%
\theoremstyle{thmstyletwo}%

\theoremstyle{thmstylethree}%

\begin{document}

\title[Complex Networks for Pattern-Based Data Classification]{Complex Networks for Pattern-Based Data Classification}


\author[1]{\fnm{Josimar} \sur{Chire}}\email{jecs89@usp.br}

\author*[2]{\fnm{Khalid} \sur{Mahmood}}\email{khalid.mahmood@it.uu.se}

\author[3]{\fnm{Zhao} \sur{Liang}}\email{zhao@usp.br}

\affil*[1]{\orgdiv{Institute of Mathematics and Computer Science}, \orgname{University of São Paulo}, \orgaddress{\city{São Carlos}, \state{SP}, \country{Brazil}}}

\affil[2]{\orgdiv{Department of Information Technology}, \orgname{Uppsala University}, \orgaddress{\city{Uppsala}, \country{Sweden}}}

\affil[3]{\orgdiv{Department of Computing and Mathematics}, \orgname{University of Sao Paulo}, \orgaddress{\city{Ribeirão Preto}, \state{SP}, \country{Brazil}}}


\abstract{Data classification techniques partition the data or feature space into smaller sub-spaces, each corresponding to a specific class. To classify into subspaces, physical features e.g.,  distance and distributions are utilized. This approach is challenging for the characterization of complex patterns that are embedded in the dataset.  However, complex networks remain a powerful technique for capturing internal relationships and class structures, enabling High-Level Classification. Although several complex network-based classification techniques have been proposed, high-level classification by leveraging pattern formation to classify data has not been utilized. In this work, we present two network-based classification techniques utilizing unique measures derived from the Minimum Spanning Tree and Single Source Shortest Path. These network measures are evaluated from the data patterns represented by the inherent network constructed from each class. We have applied our proposed techniques to several data classification scenarios including synthetic and real-world datasets. Compared to the existing classic high-level and machine-learning classification techniques, we have observed promising numerical results for our proposed approaches. Furthermore, the proposed models demonstrate the following distinguished features in comparison to the previous high-level classification techniques: (1) A single network measure is introduced to characterize the data pattern, eliminating the need to determine weight parameters among network measures. Therefore, the model is largely simplified, while obtaining better classification results. (2) The metrics proposed are sensitive and used for classification with competitive results.
}




\maketitle

\section{Introduction}

Data classification maps the training dataset to the corresponding desired output. This constructed map - called a classifier, is used to predict the output for the new input instances.  One of the primary challenges of data classification is the feature extraction. In the context of machine learning, the feature extraction is a process of transforming raw data into a set of measurable properties or characteristics.  This process involves selecting and/or creating new variables that encapsulate the essential information needed to perform a specific analysis or task. The process often reducing the dimensionality of the data while preserving its most important characteristics. The feature extraction process is challenging because good features primarily vary from one dataset to another. 

In recent years, deep learning techniques have often been used for feature extractions, which revolutionized the classification tasks for numerous application scenarios such as object detection \cite{Redmon2016}, \cite{Ren2015}, machine translation \cite{Luong2015, Wu2016}, and speech recognition \cite{Hinton2012}. Deep learning models are ideal for neural networks that represent hierarchical structures, where simpler patterns are combined and reused to form more complex patterns. The primary advantage is that they can extract suitable features from the original data in an automatic manner.  Further, classic deep learning models, such as Convolutional Neural Networks (CNN) \cite{Goodfellow2016} are often suitable for processing data having regular forms (e.g., images).  However, the high-level semantic features embedded in datasets are difficult to decompose using traditional deep-learning techniques. This types of task requires analyzing the input data as a whole to identify the relationships among data samples and, consequently, the formation of its global pattern. This situation emerges in many real-world applications, such as machine translation and medical image diagnosis. 

To leverage the pattern from the dataset, a class of deep neural networks, called Graph Neural Networks (GNNs) \cite{Ward2022, Zhang2018, Wu2019, Zhou2022} has garnered significant attention. GNNs process data represented as graphs where the global structure of the data is captured.  Although several initiatives \cite{Angelov2020, Bai2021, Ras2022, Belle2021} have been conducted, deep learning models including GNNs are still short of a mechanism to provide an effective outcome, which is particularly important for many real-world applications (e.g., medical diagnostics).

In recent years, interest in complex networks (i.e., a large-scale graph with nontrivial connection patterns) has grown considerably \cite{Barabasi99, barabasi2016network, newman10, chiresaire2020new}. This rise in interest is due to the inherent advantages of representing data as networks, which allow for capturing spatial, topological, dynamical, and functional relationships within large datasets. As a result, complex networks provide an effective method for identifying data patterns by considering the local, intermediate, and global relationships among data samples. Promising results have already been achieved in this area \cite{silva2016}.

Another approach – called hybrid classification technique \cite{silva2012} that combines both low and high levels of learning, has been proposed. Low-level classification techniques capture the physical features (e.g., geometrical or statistical features) of the input data using traditional classification methods. In contrast, high-level classification techniques utilize the complex topological properties of networks constructed from the input data. This approach typically uses three network measures—average degree, clustering coefficient, and assortativity—to represent the pattern of each class network derived from the input data. The authors \cite{silva2015} has also introduced a network-based classification technique that uses the average lengths of the transition and the attractive cycle of the tourist walk initiated from each node to represent network patterns. Another network-based classification technique employing the community concept has been proposed for detecting stock market trends \cite{TiagoZhao2021}.

The complex network-based classification approach proposed in these works present definite advantages, such as classification according to pattern formation of the data, the classification process, and interoperability. However, these works exhibit the following limitations: 
\begin{enumerate}
  \item These methods require collaboration with a traditional classification technique, creating a hybrid approach. It introduces the additional challenge of determining the weights between the two classifiers for different classification scenarios.


 \item Several network measures are utilized to characterize the data patterns represented by the constructed networks for each class. As with the previous issue, defining the weights among these measures is non-trivial.


  \item The classification process involves checking how well the new data conforms to the pattern of each class network. Since only one test data point is inserted at a time, the variations in network measures before and after the insertion are usually minimal, making it difficult to assess conformance levels.

\end{enumerate}

\subsection{Contributions}
To overcome the above-mentioned problems, we present two network-based classification techniques considering a unique measure extracted from the inherent pattern of the data represented as a graph. These two techniques: Minimum Spanning Tree (MST) \cite{MST} and Single Source Shortest Path (SSSP) \cite{SSSP}, are utilized to characterize the networks constructed for each class. These approaches eliminate the need to determine any weights in the new model, making the new measure highly sensitive to the addition of even a single data item. Our observations indicate that these techniques produce promising numerical results when compared to traditional and other high-level classification techniques.


In summary, the contributions of this work are as follows:

\begin{itemize}
  \item We have present network-based classification techniques utilizing Minimum Spanning Tree (MST) and Single Source Shortest Path (SSSP) by leveraging the data represented as a graph. While our earlier work \cite{paperjecs} has introduced the MST model, in this work we provided extensive experiments to bolster our novel approach. The improved SSSP approach on the other hand is entirely novel and has not been proposed earlier. The techniques are presented in Section \ref{methods}.
  
  \item We have provided the running time of the implementation of our proposal using MST and SSSP approach, in Section \ref{complexity}. We have shown in practice that the SSSP approach will run faster compared with MST. By performing experimental evaluation in Section \ref{analysis}, we have also verified our claim that SSSP provides better performance.


\item By utilizing synthetic and traditional datasets (e.g., Iris \cite{iris}, Wine \cite{wine}), we have demonstrated that both MST and SSSP provide comparable performance for the insertion of elements for both the same and different classes (Section \ref{experiment}).

  
  \item We have further compared our approaches to the traditional machine learning algorithms by utilizing three real-world application datasets \cite{penguin, pulsar, covid} in Section \ref{applicaiotn}. We have shown that both MST and SSSP measures provide comparable performance to the machine learning algorithms such as MLP \cite{MLP}, XGBoost \cite{XGBoost}, Gaussian Naive Bayes \cite{naive}, Multilayer Perceptron (MLP) \cite{MLP}, Decision Tree\cite{decision_tree},  Logistic Regression \cite{logistic_regression}, Gaussian Naive Bayes \cite{naive}, Gradient Boosting \cite{gradient_descent}, Bootstrap Aggregating  \cite{bagging}, and Xgboost \cite{XGBoost}, while leveraging the internal structure of the network.

\end{itemize}

\section{Methodologies} \label{methods}


The training and classification process utilizes the Minimum Spanning Tree (MST) or Single Source Shortest Path (SSSP) as a network measure. The proposed approaches consist of the following steps,  which are further elaborated through Figure \ref{fig:proposal}:

\begin{figure*}[!hbpt]
\centerline{
\includegraphics[width=0.95\textwidth]{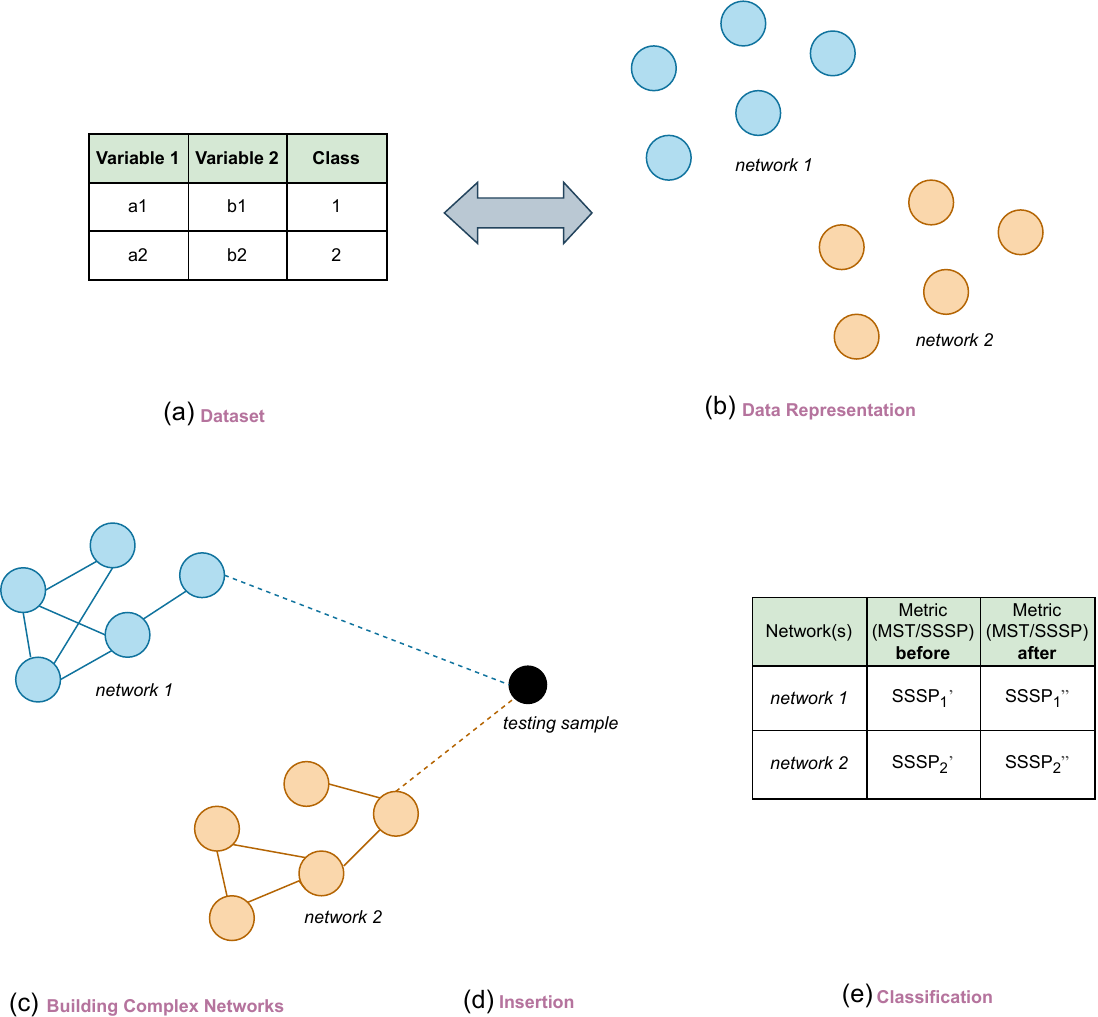}
}
\vspace*{6mm}
\caption{Proposed approaches}
\label{fig:proposal}
\end{figure*}

\begin{enumerate}
  \item In the training phase, a set of $K$ networks are constructed, each for one of the $K$ classes. This step is depicted in Fig. \ref{fig:proposal}a, where the \textit{Dataset} is classified into two \textit{classes} (i.e., class \textit{1} and \textit{2}).  
  \item For each network, a data sample is represented as a node, where the connection weight between a pair of nodes is determined by the Euclidean distance. The corresponding underlying network of the \textit{Dataset} (from Fig. \ref{fig:proposal}a) is presented in Fig. \ref{fig:proposal}b. Here, the classification marked by \textit{class 1} and \textit{class 2} corresponds to \textit{network1} and \textit{network2} respectively.
  \item The training and classification process utilized either MST or SSSP as network measures. The MST and SSSP are applied to the underlying networks of \ref{fig:proposal}b (\textit{e.g., network1}, \textit{network2}), and the corresponding connected networks (forms a tree) are shown in Fig. \ref{fig:proposal}c.
     \begin{itemize}
         \item \textbf{MST}: For each network, Minimum Spanning Tree is calculated to represent the pattern formation of the corresponding class of data. 
         \item \textbf{SSSP}: The Single Source Shortest Path algorithm requires a source to be present in the network. To select a candidate source,  we first calculate the centroid \cite{Centroid} for each network and utilize it as a source for SSSP algorithm. In this approach, we use SSSP measure (instead of MST) to represent the pattern formation.
     \end{itemize}
  \item In the classification phase, a \textit{testing sample} (shown in Fig. \ref{fig:proposal}d) is inserted into each of the $K$ networks (of Fig. \ref{fig:proposal}c). The chosen measure (either MST or SSSP) is calculated again by considering the insertion of this new \textit{testing sample}. This process is depicted in Fig. \ref{fig:proposal}e, where SSSP is chosen as an example the network metric. For \textit{network1}, the SSSP measure before inserting the \textit{test data} is \textit{SSSP\textsubscript{1}'}, while \textit{SSSP\textsubscript{1}"} is evaluated after the insertion.
  \item Finally, the \textit{testing sample} is classified into a class, where its insertion causes the smallest variation of the MST or SSSP measure. For example, if the relative variation of \textit{network1} (based on  \textit{SSSP\textsubscript{1}'} and \textit{SSSP\textsubscript{1}"}) is smaller than the variation of \textit{network2} (based on  \textit{SSSP\textsubscript{2}'} and \textit{SSSP\textsubscript{2}"}), the \textit{testing sample} will be classified as \textit{network1}. This classification process is further discussed in Section \ref{classification}.
\end{enumerate}

In these approaches, the \textit{testing sample} aligns with the pattern formation of its class. Notice that the \textit{testing sample} can be either close to or far from the training samples of the same class, as classification is based on pattern conformation. Therefore, in the proposed approaches, checking physical distance or distribution is not a criterion for classification.

It is important to note that all the steps (from 1 to 5) in the classification and training processes for both MST and SSSP are identical, with the only difference being the selection of either MST or SSSP algorithm as the network measure.

The technicalities of the proposed models are further described in the following subsections.

\subsection{Evaluation of the Network Structure}
High-level classification techniques can employ traditional complex network metrics, such as average degree, clustering coefficient, and assortativity, to characterize the data patterns of each class. During classification, only one test data item is added to the network, which is typically large. Consequently, this network measures reflect only minor and localized changes, making it challenging to assess the pattern conformity of the test data sample. Hence, a network-sensitive measure is needed. Therefore, we propose new network measures based on Minimum Spanning Tree (MST) or Single Source Shortest Path (SSSP).

\subsubsection{Minimum Spaning Tree (MST)}

An undirected graph $G = (V, E)$ consists of a set of vertices 
$V$ and edges $E$, where each edge  $(u,v) \in E$ connects two vertices, $u$ and $v$. Each edge has an associated weight 
$w(u,v)$. A Minimum Spanning Tree (MST) is a subset of the edges that connects all the vertices with the minimum total weight, without forming any cycles \cite{cormenbook}.

\begin{equation}
    w(G) = \sum_{(u,v) \in V} w(u,v)
\end{equation}


The MST provides the shortest path to minimally connect all nodes. Many edges from the original graph may not appear in the MST. Additionally, the MST metric can capture the network structure of each class and demonstrate significant changes before and after introducing a test sample.

\subsubsection{Single Source Shortest Path (SSSP)}
An undirected (also directed) graph $G = (V, E)$, where $V$ are the vertexes. Two vertexes, $v_1$ and $v_2$ are connected by edge $E_{(1,2)}$  and there is a associated cost for each edge $w_{(1,2)}$. A particular vertex, $v_s$ is defined as a source vertex.

The shortest path from source $v_s$ to a destination vertext $v_d$ is the path $P_(s,d) = (v_1, v_2, ..., v_n)$, where $v1 = v_s$ and $v_n = v_d$ over all possible $n$, minimizes the sum:
\begin{equation}
    P_{s,d} = \sum_{(i,j) \in n, \;  i \neq j} w(u,v)
\end{equation}

In another words, the Single Source Shortest Path from the source vertex, $v_s$ to all the other vertex, where the sum, $w(G)$ of all the paths is minimized \cite{SSSP}.

Similar to MST, SSSP forms a tree if the graph is connected, where many edges of the original network are not present in SSSP.  For each network, we do not have a defined source node, therefore, we calculated the centroid \cite{Centroid} of the graph. This centroid was defined as a source node to calculate the SSSP for each network. Like MST metric, SSSP can not only represent the structure of each class network but is also utilized to calculate the variation before and after the insertion of the testing sample.

\subsection{Implementation and Running-time of the Network Measures} \label{complexity}
In the implementation of the Minimum Spanning Tree, we have utilized Kruskal's algorithm\cite{MST}. This greedy algorithm first sorts the edges of the graph and maintains a disjoint-set data structure \cite{disjoint} to detect the cycle. If the graph $G(V,E)$ has $E$ edges, the running time to sort the edges with a comparison sort is $O(E \log {}E)$. The implementation of disjoint-set data structure uses Inverse Ackermann Function \cite{Ackermann}, which typically grows very slowly and the running time is amortized constant (i.e. $O(1)$) for each operation. Since there can be total $E$ edges that need to be checked for cycle detection using disjoint-set data structures, the running time to maintain the data structure is $O(E)$. The running time for Kruskal's Algorithm is dominated by the sorting; therefore, the complexity of the overall algorithm is $O(E \log {}E)$.

The Implementation of the Single Source Shortest Path uses Dijkstra's algorithm \cite{SSSP}. In Dijkstra's algorithm, we have used min-priority queue data structure \cite{pq} for storing and querying partial solutions sorted by weight from the source. The running time depends upon the cost of maintaining the priority queue. For an undirected graph $G = (V, E)$, 
the priority queue in our implementation can hold a maximum of $V$ edges for each vertex. Therefore, the cost of both search and insert in the priority queue is at most $O(\log {}V)$. Since we need to perform insert/serach for at most $E$ edges, the overall running time of SSSP for our implementation is $O(E \log {}V)$.

Note that, the running time of MST is $O(E \log {}E)$, while for SSSP it is $O(E \log {}V)$. However, the $O(E \log {}E)$  running time of MST can be reduced to $O(E \log {} V^2)$ for a complete graph, which is $2  \cdot O( E \log {}V)$. As a result, the MST might run slower than SSSP in practice, even though the theoretical running time of both MST and SSSP is $O( E \log {}V)$.

\subsection{Classification} \label{classification}


After constructing the network for each class, we compute an appropriate network measure to represent the pattern of each class. These values are labeled as $G_{before}(class_x)$, for $x = 1, 2, ... M$, where $M$ is the total number of classes.

In the classification phase, a test sample is introduced into each class network through the following steps:

\begin{itemize}
    \item Retrieve the adjacency matrix of the complex network for each class.
    \item Calculate the distance between the inserted test sample and all existing samples in the class.
    \item Update the adjacency matrix to include the new sample.
\end{itemize}


After the test sample is inserted, the same network measure, $G_{after}(class_x)$, for $x = 1, 2, ..., M$, is computed and compared with the original measure ( MST or SSSP) before the insertion (denoted as $G_{before}(class_x)$) for each class of the network. This comparison shows the effect of adding the new sample to each class, represented as:
\begin{equation}
\begin{split}
\Delta G (class_x) = || G_{before} (class_x) - G_{after} (class_x) ||\,, \; \\
x = 1,2, ..., M.
\end{split}
\label{eq:network_perturbation}
\end{equation}

Finally, the test sample is classified into the class $y$, where:

\begin{equation}
\Delta G (class_y) = min \{\Delta G (class_x) \}\,, \; x = 1,2, ..., M.
\label{eq:min_perturbation}
\end{equation}


Specifically, the change in the MST or SSSP metric, $w(G)$, is observed before and after the test sample is added to each class network. The classification is based on the pattern-matching rule, meaning the test sample will be assigned to the class where its insertion results in the smallest change in $w(G)$.


\section{Experimental Evaluations  } \label{experiment}
In this section, we present the experimental results by applying the MST and SSSP techniques to the synthetic and real datasets.

\subsection{Sensitivity of Metric}

Our earlier work \cite{paperjecs} provides a baseline experiment using only MST to test the sensitivity of traditional metrics like clustering coefficient,  assortativity. It demonstrated that MST is sensitive to changes in the network structure.  In this work, the experiments are further extended to provide a comparable view of sensitivity for both MST and SSSP.  This will help us to understand how sensitive SSSP is to the change of the network structure compared to the MST and other measures.  

In our experiments, we have utilized two datasets: (1) an artificial dataset using normal distribution and (2) iris \cite{iris} and wine \cite{wine} dataset. 


\subsubsection{Artificial dataset with normal distribution}

The samples are generated using the normal distribution, by utilizing the following Eq. \ref{normal_distribution}.

\begin{equation}
    f(x) = \frac{1}{\sigma \sqrt{2\pi}} \exp\left(-\frac{(x - \mu)^2}{2\sigma^2}\right)
\label{normal_distribution}
\end{equation}
where:
\begin{align*}
f(x) & \text{ is the probability density function.} \\
x & \text{ is the value of the random variable.} \\
\mu & \text{ is the mean of the distribution.} \\
\sigma & \text{ is the standard deviation of the distribution.}
\end{align*}

\begin{figure}[H]
  \centerline{
  \includegraphics[width = 0.6\columnwidth]{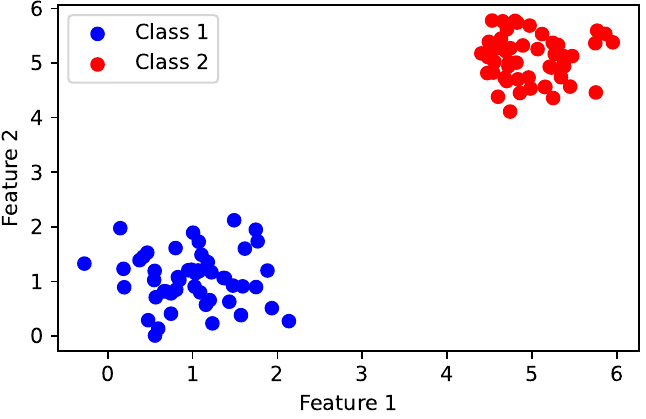}
  }
\caption{Synthetic dataset following a Normal Distribution with two classes, 50 samples}
\label{fig:4_twoblobs}
\end{figure}

A generated dataset of two classes with normal distribution each is shown by Fig. \ref{fig:4_twoblobs}, and the parameter values are: 

\begin{itemize}
    \item First class: $\mu$ = [1,1], $\sigma$ = [0.5, 0.5]
    \item Second class: $\mu$ = [5,5], $\sigma$ = [0.4, 0.4]
\end{itemize}

\begin{figure}[ht]
\captionsetup{justification=centering}
    \begin{subfigure}{0.45\textwidth}
        \includegraphics[width=\hsize]
        {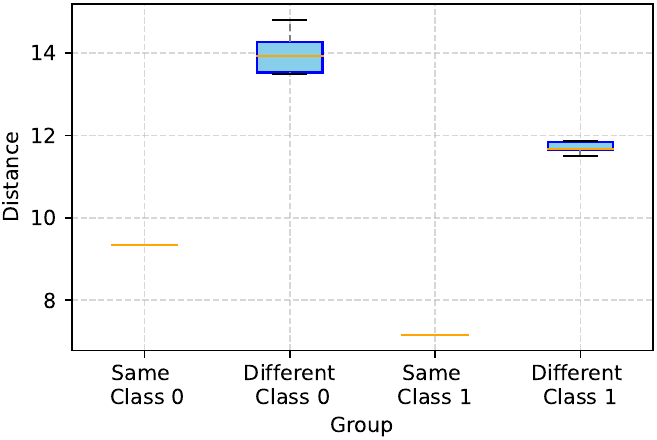}
        \caption{Boxplot of MST Distances for Insertion of Same/Different Elements}
        \label{fig:mst-normal}
    \end{subfigure}
\hfill
    \begin{subfigure}{0.45\textwidth}
    \includegraphics[width=\hsize]
    {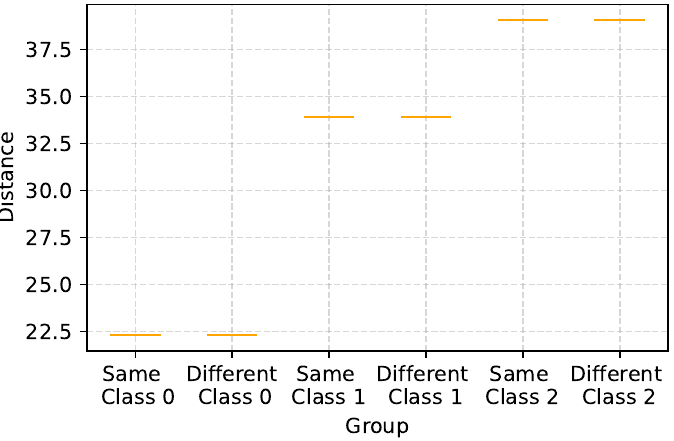}
        \caption{Boxplot of SSSP Distances for Insertion of Same/Different Elements}
    \label{fig:sssp-normal}
\end{subfigure}
    \caption{Sensitivity experiment for MST and SSSP using synthetic dataset generated with Normal Distribution}
    \label{fig:4_artdat_mst}
\end{figure}

The result of the the insertion of 5 samples belonging to the same and different classes, is presented in Fig. \ref{fig:4_artdat_mst}. This provides insight into the performance of the proposal using MST (Fig. \ref{fig:mst-normal}) and SSSP (Fig. \ref{fig:sssp-normal}), indicating that sensitivity is required for future classification tasks.

The findings from this analysis allow us to evaluate how sensitive network metrics are when new elements are introduced to the class structure.


\pagebreak
\subsection{Iris and Wine dataset} 

\begin{figure}[ht]
\captionsetup{justification=centering}
    \begin{subfigure}{0.45\textwidth}
        \includegraphics[width=\hsize]
        {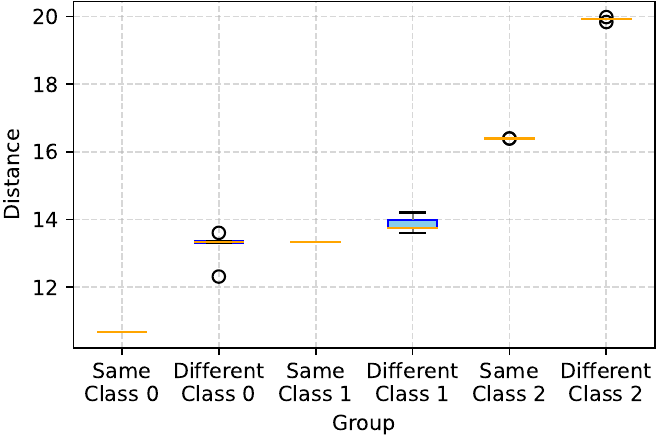}
        \caption{Boxplot of MST Distances for Insertion of Same/Different Elements}
        \label{fig:mst-iris}
    \end{subfigure}
\hfill
    \begin{subfigure}{0.45\textwidth}
    \includegraphics[width=\hsize]
    {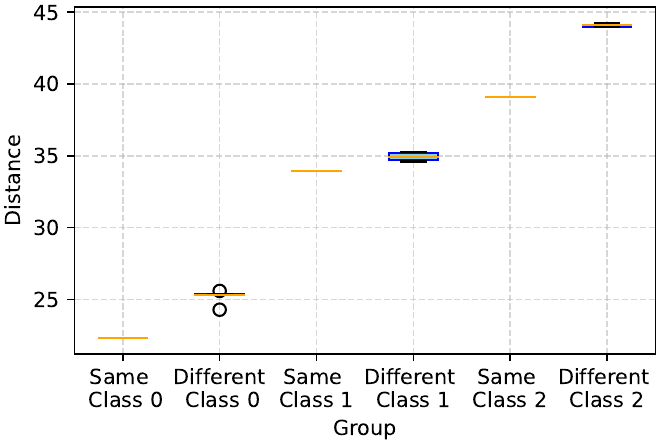}
        \caption{Boxplot of SSSP Distances for Insertion of Same/Different Elements}
    \label{fig:sssp-iris}
\end{subfigure}
    \caption{Sensitivity experiment for MST and SSSP using Iris Dataset}
    \label{fig:4_iris_mst}
\end{figure}

\begin{figure}[ht]
\captionsetup{justification=centering}
    \begin{subfigure}{0.45\textwidth}
        \includegraphics[width=\hsize]
        {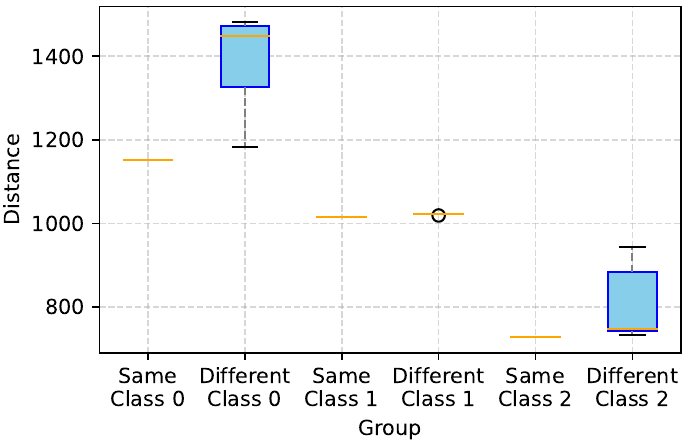}
        \caption{Boxplot of MST Distances for Insertion of Same/Different Elements}
        \label{fig:mst-wine}
    \end{subfigure}
\hfill
    \begin{subfigure}{0.45\textwidth}
    \includegraphics[width=\hsize]
    {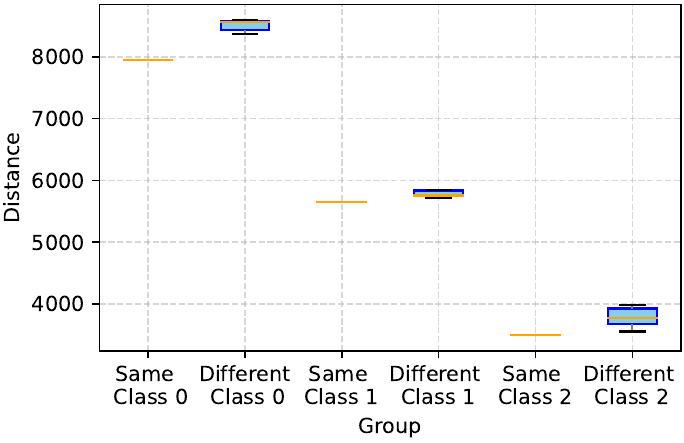}
        \caption{Boxplot of SSSP Distances for Insertion of Same/Different Elements}
    \label{fig:sssp-wine}
\end{subfigure}
    \caption{Sensitivity experiment for MST and SSSP using Wine Dataset}
    \label{fig:4_wine_mst}
\end{figure}

In this section, the experimental results of the sensitivity measures using the MST and SSSP metrics utilizing two real dataset are performed. The results of the Iris dataset \cite{iris} are depicted in Fig. \ref{fig:4_iris_mst}, while the Wine dataset \cite{wine} is depicted in Fig. \ref{fig:4_wine_mst}. Upon analyzing these results, it is evident that both MST (Fig. \ref{fig:mst-iris} and Fig. \ref{fig:mst-wine}) and SSSP (Fig. \ref{fig:sssp-iris} and Fig. \ref{fig:sssp-wine}) measures are highly sensitive to the insertion of a test sample into a class to which it does not belong. In this scenario, the variation in the MST and SSSP measures are moderately significant. Conversely, when a test sample is inserted into the class to which it belongs, the variation is consistently small.

Due to its high sensitivity and efficiency, both MST and SSSP measures are used to perform all the classification tasks in this work.

\subsection{Performace Comparison between MST and SSSP} \label{analysis}
We have conducted 1000 experiments, considering three complex networks for both MST and SSSP measures. The execution times of the result are provided in Table \ref{results}. We have observed that the \textit{mean} execution time for MST is 2.891 milliseconds(ms) having a standard deviation of 0.633 ms (\textit{std. dev.} in the table). For SSSP, the \textit{mean} execution time is 1.131 ms, with a standard deviation of 0.449 ms. The minimum (\textit{min}), maximum (\textit{min}), and different percentiles (\textit{25\%, 50\%, \& 75\%}) are also provided in Table \ref{results}. 
 
\begin{table}[hbpt]
\caption{Performance comparison between MST and SSSP. \\ 
\label{results}}
\centering
\begin{tabular}{|r|c|c|}
\hline
              & \textbf{MST} & \textbf{SSSP} \\ 
              & \textit{time(ms)}     & \textit{time(ms)}    \\ \hline
\textbf{mean} & 2.891     & 1.131     \\ \hline
\textbf{std. dev. }  & 0.633     & 0.449     \\ \hline
\textbf{min}  & 2.314     & 0.819     \\ \hline
\textbf{max}  & 19.585     & 25.247     \\ \hline
\textbf{25\%} & 2.447     & 0.901     \\ \hline
\textbf{50\%} & 2.689     & 0.983     \\ \hline
\textbf{75\%} & 3.190     & 1.152     \\ \hline
\end{tabular}
\end{table}

The most important observation of these experiments is that the mean execution time for SSSP is approximately 2.5 times faster than the performance of MST. This significant performance difference is attributed to the different approaches of the implementation of the algorithms. Our implementation of MST uses Kruskal's algorithm, which sorts the entire network first, while SSSP uses Dijkstra's algorithm maintains a priority queue stroing the partial network. As a result, the sorting process in MST significantly impacts its overall execution time. As discuss in the Section \ref{complexity}, the actual running time of SSSP is $O( E \log {}V)$, while the running time of MST is $2  \cdot O( E \log {}V)$. Therefore, the experimental performance of 2.5 speed up of the SSSP compared with MST matches with running time of the implementations provided in Section \ref{complexity}.

\pagebreak
\section{Performance Evaluation of Real-world Applications} \label{applicaiotn}

This section provides the performance evaluations of our network-based classification techniques compared with the traditional machine learning algorithms by utilizing three real-world application datasets. We have compared both the MST and SSSP measures with machine learning algorithms such as  Multilayer Perceptron (MLP) \cite{MLP}, Decision Tree\cite{decision_tree},  Logistic Regression \cite{logistic_regression}, Gaussian Naive Bayes \cite{naive}, Gradient Boosting
  \cite{gradient_descent}, Bootstrap Aggregating (a.k.a Bagging) \cite{bagging}, and Xgboost \cite{XGBoost}. 

The datasets were drawn from the real-world applications utilizing the Penguine Classification dataset \cite{penguin}, Pulser Star Detection Classification dataset \cite{pulsar}, and the COVID-19 Computed Tomography (CT) scan classification dataset \cite{covid}. For the experiment, we have used a cross-validation of K=10.

\subsection{Penguin Classification}

The Palmer Archipelago (Antarctica) Penguin Dataset \cite{penguin} was collected and made available by Dr. Kristen Gorman and the Palmer Station, Antarctica (a member of the Long Term Ecological Research Network). 
The dataset is widely used for ecological and biological research and includes detailed measurements and observations on three species of penguins: Adélie, Chinstrap, and Gentoo. The dataset consists of $344$ observations having $8$ variables. Each observation represents a single penguin, and the variables are described as follows:
\begin{table}[htbp]
\caption{Descriptions of the Penguin dataset.} \label{penguin}
\centering
\small
\begin{tabular}{|r|l|}
\hline
\textbf{\normalsize{Variable}} & \textbf{\normalsize{Description}} \\ \hline
\textbf{species} & The species of the penguin (\textit{Adelie}, \textit{Chinstrap}, or \textit{Gentoo}) \\  \hline
\textbf{island} & Location of the penguin (\textit{Biscoe}, \textit{Dream}, or \textit{Torgersen}) \\  \hline
\textbf{bill\_length\_mm} & The length of the penguin's bill in millimeters \\  \hline
\textbf{bill\_depth\_mm} & The depth of the penguin's bill in millimeters \\  \hline
\textbf{flipper\_length\_mm} & The length of the penguin's flipper in millimeters \\  \hline
\textbf{body\_mass\_g} & The body mass of the penguin in grams \\  \hline
\textbf{sex} & The sex of the penguin (\textit{male} or \textit{female}) \\  \hline
\textbf{year} & The year the observation was recorded (2007 or 2008) \\  \hline
\end{tabular}
\end{table}

\begin{figure}[hbpt]
  \centerline{
  \includegraphics[width = 1.0\columnwidth]{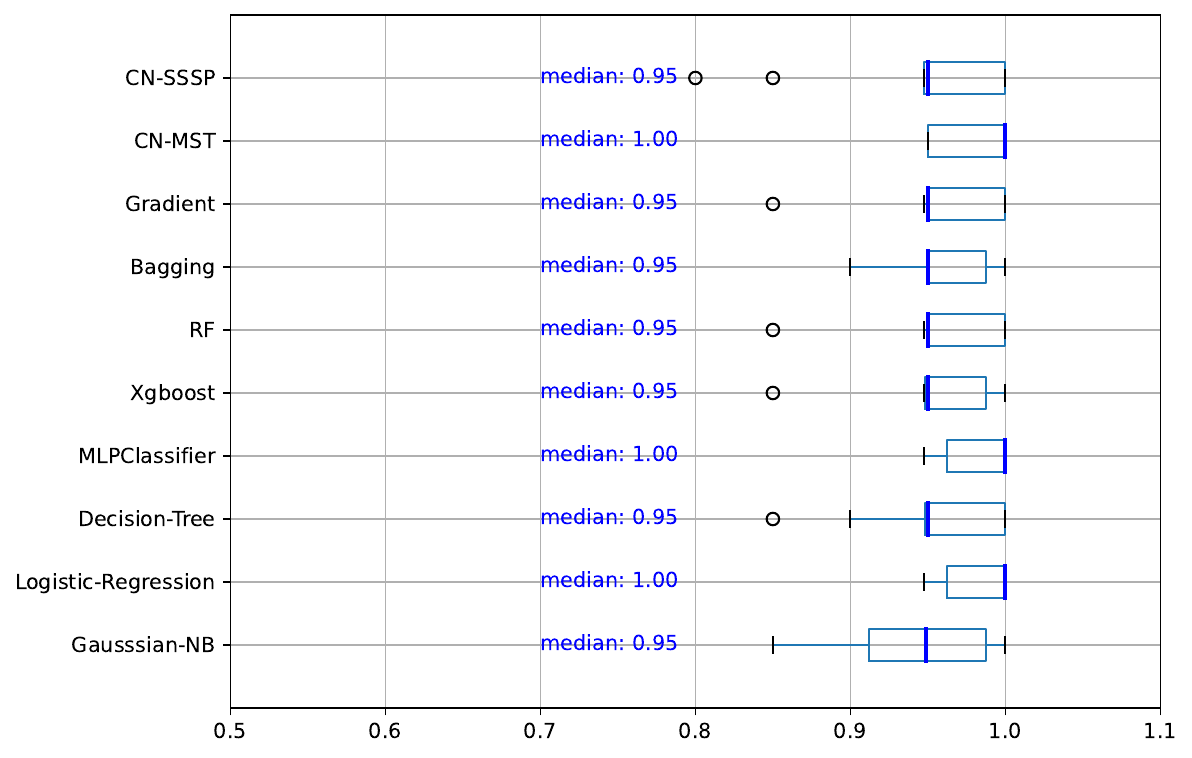}
  }
\caption{Classification accuracy results of  Penguin dataset.}
\label{fig:penguin}
\end{figure}

%
We have classified the data for the three classes for each species using MST, SSSP, and different machine learning techniques. The results of the classifications using these approaches are depicted in Fig. \ref{fig:penguin}. 

From the result, it is evident that for all the algorithms the accuracy median is equal to or higher than 0.95; therefore, all algorithms perform well for the Penguin dataset. The accuracy median of SSSP provides comparable performance having a comparable standard deviation to the ML algorithms (e.g., Xgboost, Gaussian-NB), and MST approaches. Furthermore, the MST provides equivalent performance to traditional machine learning algorithms.




\subsection{Pulsar Star Detection}
Pulsars are rotating neutron stars, characterized by intense magnetic fields, which are swiftly spinning to emit electromagnetic radiation in concentrated beams. These emissions are displayed as recurring pulses across different wavelengths. Identifying pulsars from large amounts of data has significant importance in the field of radio astronomy. The High Time Resolution Universe survey dataset, version 2 (HTRU2) \cite{pulsar}, consists of collection of pulsar candidate and non-pulsar candidate examples. The dataset was contributed to the University of California, Irvine's machine learning repository by Dr. Robert Lyon et al. of The University of Manchester. 

The dataset consists of 17,898 examples, including both pulsar and non-pulsar instances. The dataset consists of 9 features that can be classified as two classes: 

\begin{table}[htbp]
\caption{Descriptions of the Pulsar Dataset(HTRU2). \\\label{htru}}
\centering
\small
\begin{tabular}{|r|l|}
\hline
\textbf{\normalsize{Variable}} & \textbf{\normalsize{Description}} \\ \hline
\textbf{Mean Profile} & The mean of the integrated pulse profile \\  \hline
\textbf{Standard Deviation Profile} &  Standard deviation of integrated pulse profile \\  \hline
\textbf{Excess Kurtosis Profile} & Excess kurtosis of integrated pulse profile \\  \hline
\textbf{Skewness Profile} & The skewness of the integrated pulse profile. \\  \hline
\textbf{Mean Curve} & The mean of the DM-SNR curve. \\  \hline
\textbf{Standard Deviation Curve} & The standard deviation of the DM-SNR curve. \\  \hline
\textbf{Excess Kurtosis Curve} & The excess kurtosis of the DM-SNR curve. \\  \hline
\textbf{Skewness Curve} & The skewness of the DM-SNR curve. \\  \hline
\textbf{Class Label} & Variable indicating class of a pulsar. \\  \hline
\end{tabular}
\end{table}

\pagebreak

\begin{figure}[hbpt]
  \centerline{
  \includegraphics[width = 1.0\columnwidth]{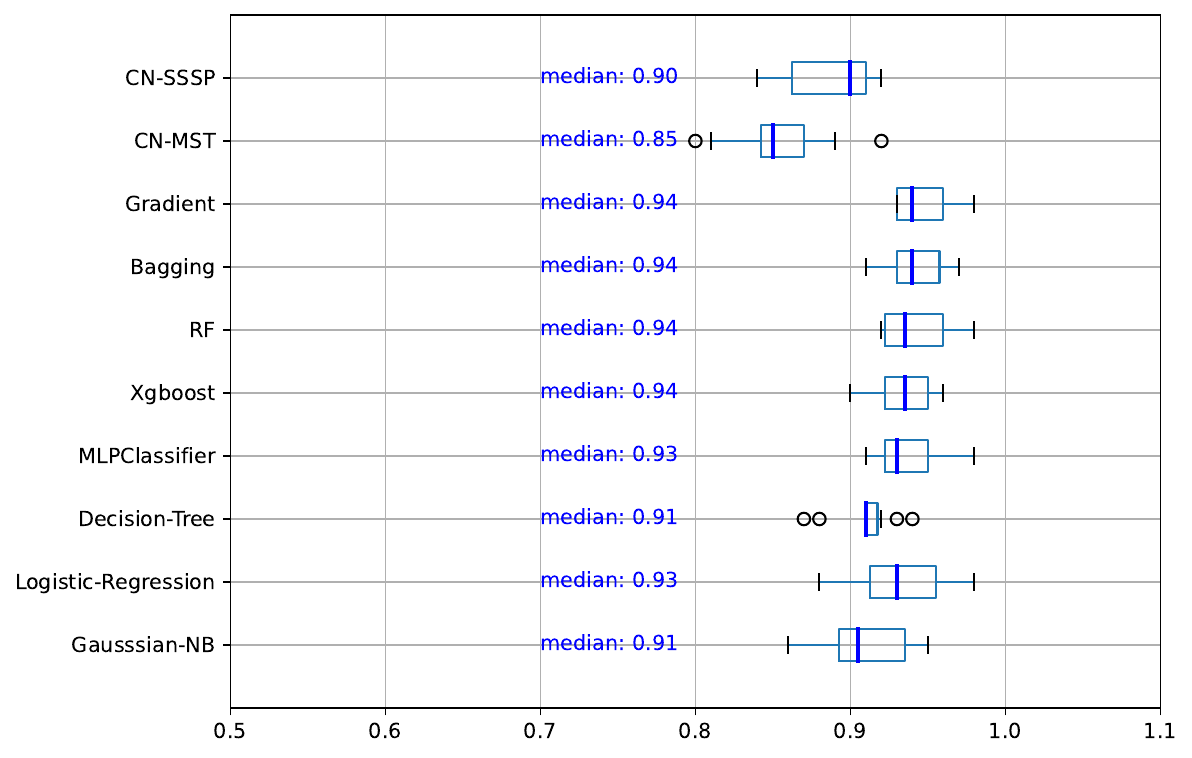}
  }
\caption{Classification accuracy results of Pulsar Star Dataset(HTRU2).}
\label{fig:HTRU}
\end{figure}

The dataset is commonly used for binary classification tasks to distinguish between pulsar and non-pulsar examples.  Novel algorithms from the fields of machine learning and astronomy utilize this dataset to evaluate its performance. The result of the classification of the HTRU2 dataset using MST, SSSP, and different machine learning techniques are depicted in Fig. \ref{fig:HTRU}.



From the result, it is evident that the accuracy median of all the algorithms has a accuracy median of 0.9 or higher except MST. While MST still has a good accuracy of 0.85, SSSP still performs better with a precison median of 0.9. Furthermore, SSSP provides comparable performance to other machine learning algorithms (e.g., Xgboost, Gaussian-NB).


\subsection{Covid-19 Classification}

Medical imagine techniques such as Chest X-ray and computed tomography scan (CT-scan) are important methods for the diagnosis of pulmonary diseases such as COVID-19. While the results can be interpreted and classified by the medical personal to identify the diseases, automated classification problems can also be effectively utilized without human intervention. In this work, we have utilized real-world COVID-19 CT-Scan images \cite{covid} for the proposed network-based classification techniques to classify the diseases.

\begin{figure}[hbpt]
  \centerline{
  \includegraphics[width = 1.0\columnwidth]{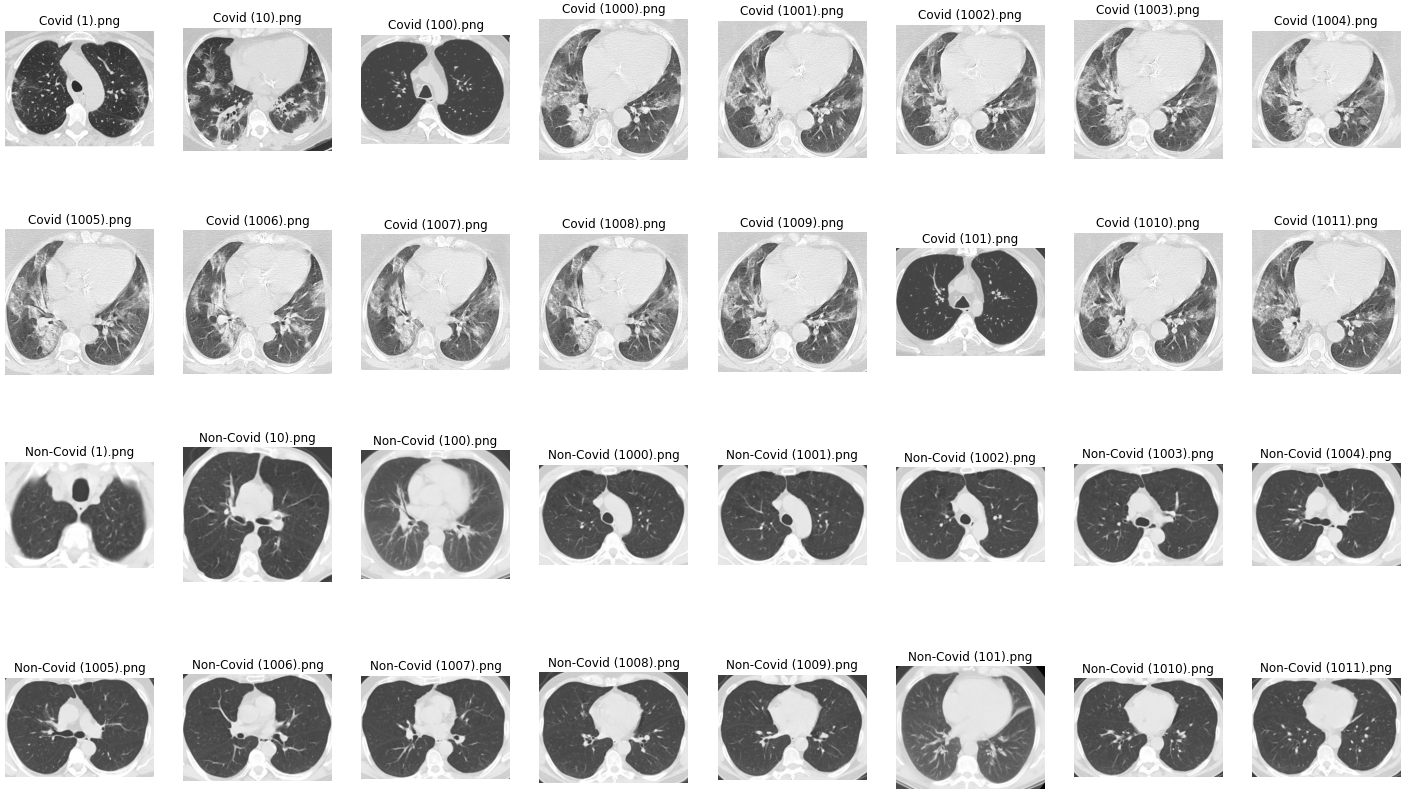}
  }
\caption{Dataset samples of SARS-COV-2 Ct-Scan Dataset \cite{covid}.}
\label{fig:dataset1}
\end{figure}

The dataset consists of  50 images for each class, which made the dataset balanced. For illustrative purposes, 16 samples for each positive and negative case are presented in Fig. \ref{fig:dataset1}.

Feature extraction is carried out using the Gray Level Co-occurrence Matrix (GLCM) to obtain image patterns \cite{Mall2019}, \cite{Singh2017}. A total of $40$ features based on GLCM are extracted, and two classes are considered.

\begin{figure}[hbpt]
  \centerline{
  \includegraphics[width = 1.0\columnwidth]{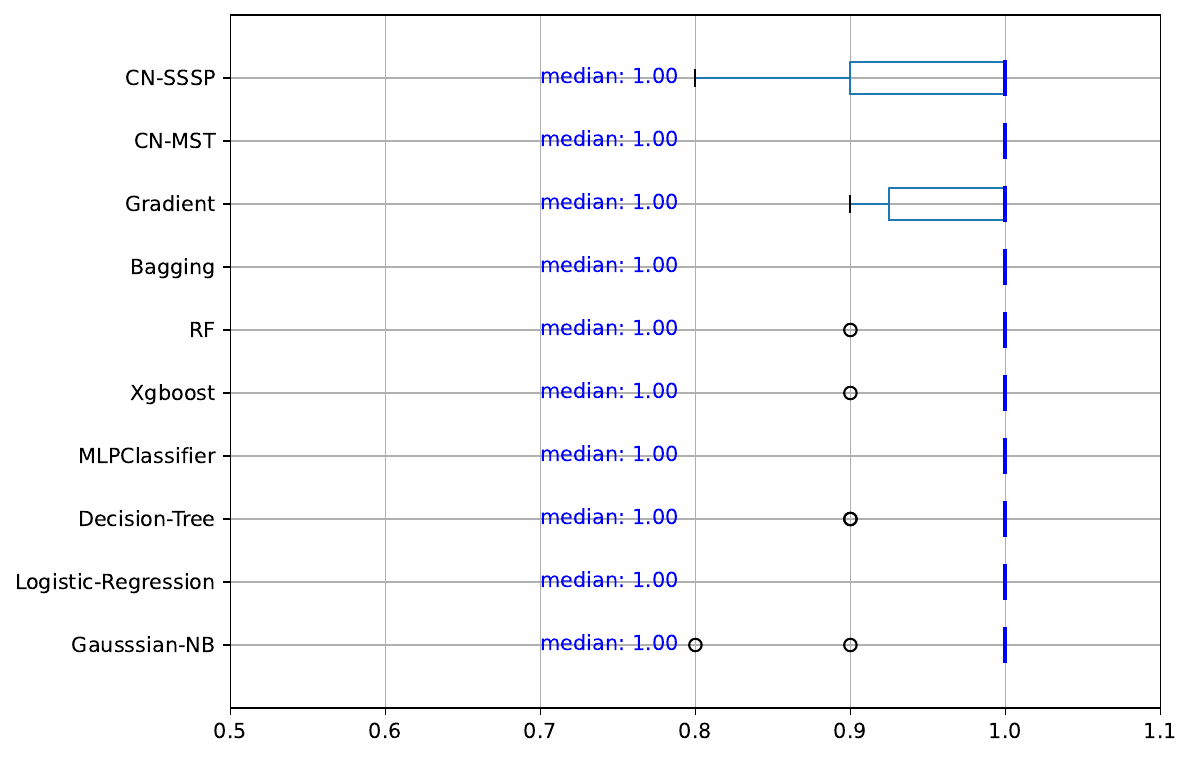}
  }
\caption{Classification accuracy results of Covid-19 dataset.}
\label{fig:covid}
\end{figure}

The result of the classification of the dataset using MST, SSSP, and different machine learning techniques are depicted in Fig. \ref{fig:covid}. 

\pagebreak



It is evident from the result that the accuracy median of SSSP has a comparable performance with the machine learning algorithms (e.g., Xgboost, Gaussian-NB, etc). Similarly, the MST also demonstrates competitive performance compared to the machine learning algorithms.








\section{Conclusion}
In this work, we presented two distinct network-based classification techniques using the Minimum Spanning Tree (MST) and Single Source Shortest Path (SSSP). Both techniques describe the data patterns represented by the network constructed for each class. 

Performance evaluations using synthetic and empirical datasets demonstrate that incorporating MST and SSP measures provides higher sensitivity to the data pattern formation, leading to improved classification outcomes. 

We also provided the execution times of implementations of our approaches. Through complexity analysis and experimental evaluation, we confirmed that the SSSP method outperforms the MST approach in terms of performance. For some datasets, the accuracy of SSSP was also observed better compared to MST for accuracy; therefore, SSSP is demonstrated to be a more competitive algorithm than MST.

Finally, we applied the proposed techniques to datasets from three real-world application scenarios. The algorithms have demonstrated comparable performance with the contemporary machine learning classification algorithms having the enhanced features of capturing complex network attributes.


\section{Future Works}
In future work, we plan to develop advanced classification techniques utilizing dynamic network measures based on the maximal flow of the underlying network. We believe that these dynamic measures can more accurately capture data patterns, leading to an improved classification result.


\section*{Acknowledgment}
The authors of this work would like to thank the Center for Artificial Intelligence (C4AI-USP) and the support from the São Paulo Research Foundation (FAPESP grant\# 2019/07665-4) and from the IBM Corporation. The authors also would like to thank the support from the China Branch of the BRICS Institute of Future Network.

The authors also want to mention Research4Tech, an Artificial Intelligence community of Latin American Researchers for promoting Science and collaboration in Latin American countries. 

\bibliography{sn-bibliography}

\end{document}